\pdfoutput=1

\documentclass[11pt]{article}

\usepackage[]{ACL2023}

\usepackage{times}
\usepackage{latexsym}

\usepackage[T1]{fontenc}

\usepackage[utf8]{inputenc}

\usepackage{microtype}

\usepackage{inconsolata}

\usepackage[ruled, vlined, linesnumbered]{algorithm2e}
\usepackage{bm}
\usepackage{bbm}
\usepackage{amsthm}
\usepackage{amsmath}
\usepackage{mathrsfs} 
\usepackage{tabularx}
\usepackage{graphicx}
\usepackage{multirow}
\usepackage{xspace}
\usepackage{xcolor}
\usepackage{enumitem}
\usepackage{placeins}
\usepackage{soul}
\usepackage{mathtools}
\usepackage{booktabs}
\usepackage{hyperref}
\usepackage{subcaption}
\usepackage{pgfplots}
\usepackage{csquotes}
\usepackage{cleveref}
\usepackage{balance}
\usepackage{multicol}
\usepackage{makecell}
\usepackage{amssymb}
\usepackage[switch]{lineno}
\DeclareUnicodeCharacter{2212}{−}
\usepgfplotslibrary{groupplots,dateplot}
\usetikzlibrary{patterns,shapes.arrows}
\pgfplotsset{compat=newest}

\newcommand{\ours}{{RAFTS}\xspace}

\title{Retrieval Augmented Fact Verification by Synthesizing Contrastive Arguments}

\author{Zhenrui Yue \quad Huimin Zeng \quad Lanyu Shang \quad Yifan Liu \\
\textbf{Yang Zhang \quad Dong Wang} \\
University of Illinois Urbana-Champaign \\
\texttt{\{zhenrui3, huiminz3, lshang3, yifan40, yzhangnd, dwang24\}@illinois.edu} \\}

\begin{document}
\maketitle
\begin{abstract}
The rapid propagation of misinformation poses substantial risks to public interest. To combat misinformation, large language models (LLMs) are adapted to automatically verify claim credibility. Nevertheless, existing methods heavily rely on the embedded knowledge within LLMs and~/~or black-box APIs for evidence collection, leading to subpar performance with smaller LLMs or upon unreliable context. In this paper, we propose \ul{r}etrieval \ul{a}ugmented \ul{f}act verifica\ul{t}ion through the \ul{s}ynthesis of contrasting arguments (\ours). Upon input claims, \ours starts with evidence retrieval, where we design a retrieval pipeline to collect and re-rank relevant documents from verifiable sources. Then, \ours forms contrastive arguments (i.e., supporting or refuting) conditioned on the retrieved evidence. In addition, \ours leverages an embedding model to identify informative demonstrations, followed by in-context prompting to generate the prediction and explanation. Our method effectively retrieves relevant documents as evidence and evaluates arguments from varying perspectives, incorporating nuanced information for fine-grained decision-making. Combined with informative in-context examples as prior, \ours achieves significant improvements to supervised and LLM baselines without complex prompts. We demonstrate the effectiveness of our method through extensive experiments, where \ours can outperform GPT-based methods with a significantly smaller 7B LLM\footnote{Our implementation is publicly available at https://github.com/yueeeeeeee/\ours.}.
\end{abstract}

\section{Introduction}

As the scope of social media and digital forums continue to expand, increasing amount of misinformation has been observed across multiple platforms (e.g., Twitter), posing risks to public interest~\cite{chen2022combating}. Therefore, fact-checking methods are proposed to prevent the spreading of false information before it leads to severe consequences~\cite{litou2017efficient, hassan2017toward, shu2017fake}. For example, online fact-checking services (e.g., Snopes\footnote{https://www.snopes.com/}) employ professional fact-checkers to identify instances of misinformation. Nevertheless, human fact-checking involves a significant amount of manual work, proving to be less efficient confronted with the vast volume of misinformation, particularly as it evolves and spreads online~\cite{micallef2020role, nakov2021automated}.

To perform fact-checking at scale, automated methods have emerged by leveraging large language models (LLMs)~\cite{shu2022cross, yang-etal-2022-coarse, yue-etal-2023-metaadapt, choi2024fact}. For example, RARG proposes to train and align LLMs for generating faithful explanations upon detected misinformation~\cite{yue2024evidence}. Despite their effectiveness, these methods typically require extensive training data and may demonstrate performance deterioration upon domain~/~concept shifts~\cite{zhu2022memory, nan-etal-2022-improving, gu2023unsupervised, shang2024mmadapt}. Moreover, many models are unaware of external evidence~/~knowledge and must be frequently re-trained to incorporate up-to-date domain knowledge for accurate fact-verification~\cite{izacard-grave-2021-leveraging, borgeaud2022improving, yue-etal-2023-metaadapt}. 

\begin{figure*}[t]
    \centering
    \includegraphics[trim=1cm 4.3cm 1cm 4.8cm, clip, width=1.0\linewidth]{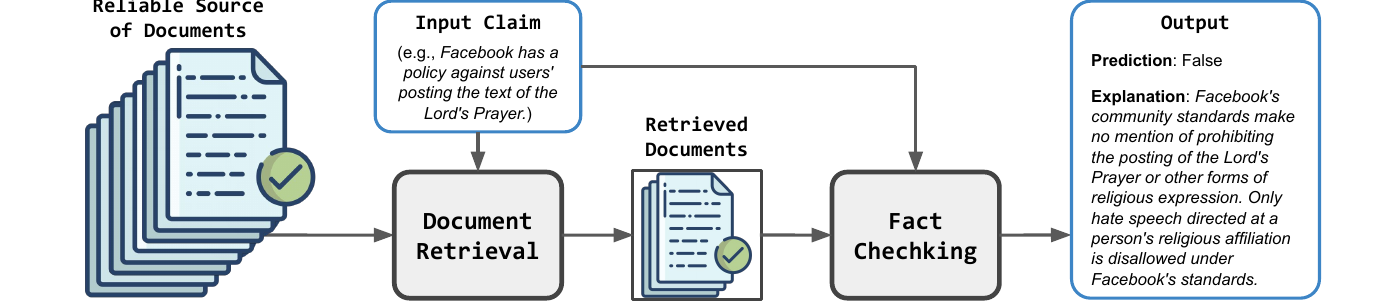}
    \caption{Our retrieval augmented generation framework for fact verification.} 
    \label{fig:intro}
\end{figure*}

As a solution, evidence-based fact-checking methods are proposed to collect evidence (e.g., documents, graphs etc.), followed by extracting relevant information and assessing the credibility of input claims through LLMs~\cite{koloski2022knowledge, kou2022crowd, shang2022privacy, wu2022adversarial, zhang-gao-2023-towards, wang-shu-2023-explainable, liu2024teller}. An example is FOLK, which leverages first-order-logic to construct sub-claims and perform question answering-based verification to generate predictions and explanations~\cite{wang-shu-2023-explainable}. Yet current approaches rely on the assumption that input claims can be decomposed into a series of predicates (i.e., sub-claims) through complex prompts. Moreover, they depend on the embedded knowledge within LLMs and~/~or black-box APIs (e.g., SerpAPI\footnote{https://www.serpapi.com/}) to collect external information, leading to subpar performance with smaller LLMs or provided with unreliable evidence~\cite{zhang-gao-2023-towards, wang-shu-2023-explainable}.

Consequently, we consider a retrieval augmented generation (RAG) framework designed to extract relevant information from reliable documents (i.e., Wikipedia, scholarly articles etc.), where the extracted information can be used as supporting facts to assess the claim credibility through LLMs. That is, given the input statement, our first objective is to retrieve (and optionally re-rank) relevant documents among an extensive collection of documents from verifiable sources. Subsequently, we utilize the retrieved documents to fact-check the input claim, aiming to either confirm the input or uncover opposing information that identifies misinformation. Our framework is visually illustrated in \Cref{fig:intro}, where the RAG-based fact verification framework retrieves relevant documents, and then generates both prediction and explanation regarding the validity of the input statement.

To this end, we propose \ul{r}etrieval \ul{a}ugmented \ul{f}act verifica\ul{t}ion through the \ul{s}ynthesis of contrastive arguments (\ours), which effectively retrieves relevant documents and performs few-shot fact verification using pretrained LLMs. \ours is structured into three components: (1)~demonstration retrieval, where informative in-context examples are collected to improve fact-checking performance; (2)~document retrieval, in which we design a retrieve and re-rank pipeline to accurately identify relevant documents for input claims; and (3)~few-shot fact verification through the synthesis of contrasting arguments. Unlike current approaches, \ours formulates supporting and opposing arguments derived from the facts within the collected documents. Combined the informative in-context examples, \ours demonstrates enhanced fact-checking performance and consistently generates high-quality explanations. To validate the effectiveness of \ours, we adopt multiple benchmark datasets and perform extensive experiments on both document retrieval and fact verification. Experiment results highlight the effectiveness of the proposed approach, where \ours can outperform state-of-the-art methods even with a significantly smaller LLM (e.g., Mistral 7B).

We summarize our contributions:
\begin{enumerate}
\item We propose a RAG-based framework, where relevant documents are retrieved from reliable sources to fact-check input claims.

\item We design \ours in three key components: demonstration retrieval, document retrieval and in-context prompting. \ours identifies informative examples and relevant documents, followed by synthesizing contrastive arguments for fine-grained fact-checking.

\item We show the effectiveness of \ours by experimenting on document retrieval and fact verification tasks. Both quantitative and qualitative results demonstrate that \ours can outperform state-of-the-art methods in fact verification and explanation generation.
\end{enumerate}
\section{Related Work}

\subsection{Large Language Models and Retrieval Augmented Generation}
Recent advancements in large language models (LLMs) have shown significantly enhanced capabilities in language comprehension and generation~\cite{raffel2020exploring, brown2020language, wei2021finetuned, ouyang2022training, chowdhery2022palm, touvron2023llama, openai2023gpt, jiang2024mixtral}. Due to the vast number of parameters and extensive quantity of pretraining corpora, LLMs can embed global knowledge within their parameters, and thus achieve significant performance improvements across diverse applications~\cite{openai2023gpt, penedo2023refinedweb, sun2023head}. However, LLMs often fail to capture fine-grained knowledge and frequently generate inaccurate or fabricated information (also known as hallucination)~\cite{peng2023check, rawte2023troubling}. To access up-to-date knowledge without costly re-training, retrieval augmented generation (RAG) has been proposed to generate text based on collected documents from verifiable sources~\cite{guu2020retrieval, lewis2020retrieval, izacard-grave-2021-leveraging, borgeaud2022improving, izacard2022few, shi2023replug, ram2023context, wang2023shall}. For example, Self-RAG can dynamically fetch external documents to generate contents through the usage of special tokens for retrieval and reflection~\cite{asai2023self}. Nevertheless, current RAG methods remain under-explored for fact verification, particularly regarding accurate evidence retrieval and fine-grained classification~\cite{wang-shu-2023-explainable, zhang-gao-2023-towards}. As such, our work studies retrieval augmented fact verification, which gathers evidence from reliable sources and integrates contrasting opinions to achieve fine-grained fact verification.

\subsection{Fact Verification and Misinformation Detection}
Fact verification methods can generally be divided into two main categories: (1)~content-based approaches, where machine learning models predict and reason over input contents (e.g., text) to identify misinformation~\cite{yue2022contrastive, jiang2022fake, yue-etal-2023-metaadapt, chen2023can, liu-etal-2023-interpretable, mendes-etal-2023-human, huang2024creation}. Incorporating additional attributes~/~modalities such as image and propagation paths can further enhance fact verification performance~\cite{shang2021multimodal, santhosh2022multi, shang2022duo, wu-etal-2022-cross, zhou2023multimodal, yao2023end, qu2024qmfnd}; (2)~evidence-based approaches, which involve gathering external knowledge (e.g., knowledge graphs or document pieces) as evidence to validate input claims and identify false information~\cite{kou2021fakesens, kou2022hc, wu2022bias, yang-etal-2022-coarse, shang2022knowledge, xu2022evidence, zhao-etal-2023-panacea, chen2023complex, wang-shu-2023-explainable, yue2024evidence, shang2024domain}. For example, HiSS adopts hierarchical step-by-step prompting with off-the-shelf LLMs and black-box question answering (QA) pipelines to perform few-shot fact verification~\cite{zhang-gao-2023-towards}. However, state-of-the-art fact verification methods primarily concentrate on improving accuracy via sophisticated prompts and~/~or intrinsic knowledge of LLMs, causing performance degrade upon smaller LLMs or domain shifts~\cite{wang-shu-2023-explainable, pelrine-etal-2023-towards, chen2023combating}. Therefore, we concentrate on retrieval augmented fact verification by collecting relevant documents from reliable sources, enabling LLMs to augment their knowledge base for claim verification. Furthermore, we exploit in-context prompting by learning from demonstrations and synthesizing contrastive arguments, and thus significantly improves fact-checking performance.
\section{Preliminary}
We consider the following problem setup: given input claim $x$ (with label $y$) and $k$-shot demonstrations $\{ (x_i, y_i) \}_{i=1}^{k}$, we aim to: (1)~retrieve a set of $m$ documents $\{ d_i \}_{i=1}^{m}$ that provide relevant information to be used as supporting evidence; and (2)~generate label $\hat{y}$ and explanation $e$ based on the input $x$, $k$-shot examples $\{ (x_i, y_i) \}_{i=1}^{k}$ and retrieved evidence $\{ d_i \}_{i=1}^{m}$. For each input $x$, we leverage a pretrained embedding model $f_{\mathrm{embed}}$ to adaptively retrieve demonstrations $\{ (x_i, y_i) \}_{i=1}^{k}$, whereas a retrieval model is learnt to predict $\{ d_i \}_{i=1}^{m}$ and provide relevant information from verifiable sources. Based on the retrieved examples and documents, the predicted $\hat{y}$ should ideally match the ground truth label $y$. In addition, the generated explanation $e$ should demonstrate desirable properties (e.g., factuality), see example in \Cref{fig:intro}. We elaborate our settings in the following.

\noindent
\textbf{Input \& Output}: Given a dataset with train and test splits $\mathcal{X}^{\mathrm{train}}$ and $\mathcal{X}^{\mathrm{test}}$, we denote the document retrieval pipeline as $f_{\mathrm{retrieve}}$ and the LLM-based fact-checking model as $f_{\mathrm{check}}$. Formally, our framework consists of two sub-problems in information retrieval (i.e., evidence collection) and fact verification (i.e., prediction and explanation), with each of the problem defined below:
\begin{itemize}[leftmargin=10pt]
    \item \emph{Document Retrieval}: Given input claim $x$, human annotated document $d$ and a collection of $n$ documents $\{ d_i \}_{i=1}^{n}$ (with $d \in \{ d_i \}_{i=1}^{n}$), our objective is to learn a retrieval model $f_{\mathrm{retrieve}}$ that ranks the claim-document pair with the highest score ($f_{\mathrm{retrieve}}(x, d) = \max \{ f_{\mathrm{retrieve}}(x, d_i) \}_{i=1}^{n}$). During training, input $x$ and $d$ can be used to learn $f_{\mathrm{retrieve}}$. In inference, we collect a subset of $m$ documents $\{ d_i \}_{i=1}^{m}$ for fact verification, where $m \ll n$.    
    \item \emph{Fact Verification}: Subsequently, we leverage both input $x$ and collected documents $\{ d_i \}_{i=1}^{m}$ from the previous step and utilize $f_{\mathrm{check}}$ to generate: (1)~prediction $\hat{y}$ on the input credibility; and (2)~explanation $e$ on the reasoning of the prediction. To perform in-context prompting, we incorporate $k$-shot examples $\{ (x_i, y_i) \}_{i=1}^{k}$ from $\mathcal{X}^{\mathrm{train}}$ as input ($x_i \neq x$). In other words, $\hat{y}, e = f_{\mathrm{check}}(\{ (x_i, y_i) \}_{i=1}^{k}, \{ d_i \}_{i=1}^{m}, x)$).
\end{itemize}

\noindent
\textbf{Learning:} Our retrieval pipeline $f_{\mathrm{retrieve}}$ is parameterized by $\theta$. To learn $f_{\mathrm{retrieve}}$, we maximize the score of the sampled input-document pair $(x, d)$. That is, we minimize the expected loss $\mathcal{L}$ over $\mathcal{X}^{\mathrm{train}}$: {$\min_{\substack{\theta}} \mathbb{E}_{(x, d) \sim \mathcal{X}^{\mathrm{train}}} [\mathcal{L}(\theta, (x, d))]$}. Meanwhile, the fact-checking model $f_{\mathrm{check}}$ (i.e., pretrained LLM) remains unchanged to minimize training expenses. To optimize fact-checking performance of $f_{\mathrm{check}}$, we employ a lightweight embedding model $f_{\mathrm{embed}}$ to select informative in-context demonstrations $\{ (x_i, y_i) \}_{i=1}^{k}$, we elaborate the details in \Cref{sec:icl-retrieval}.

\section{Methodology}

\subsection{In-Context Demonstrations}
\label{sec:icl-retrieval}
Current LLM-based approaches for fact verification utilize sophisticated prompts to identify misinformation, but depend on carefully designed prompts and \emph{static} in-context demonstrations~\cite{wei2022chain, zhang-gao-2023-towards}. Nevertheless, the classification criteria often vary from domain to domain, causing performance drops when identical prompts are applied across different contexts (as we show in \Cref{sec:exp}). In addition, diverse and informative examples are found to be helpful for performance, in particular for smaller yet more efficient LLMs~\cite{liu2021makes, zhang-etal-2022-active, levy-etal-2023-diverse, li-qiu-2023-finding}. As such, we design a retrieval pipeline to select in-context demonstrations, thereby enhancing the fact-checking performance and mitigating performance deterioration issues across domains.

We formulate the in-context learning (ICL) problem as follows. Provided with $k$-shot examples $\{ (x_i, y_i) \}_{i=1}^{k}$, we prompt a pretrained LLM with them as demonstrations to generate the fact-checking prediction $\hat{y}$ given input $x$:
\begin{equation}
    \hat{y} = \arg\max_{y} f_{\mathrm{check}}(y | \{ (x_i, y_i) \}_{i=1}^{k}, x),
\end{equation}
with $f_{\mathrm{check}}$ returning the output probabilities of the LLM. The prediction can be obtained by selecting the output with the highest probability conditioned on the provided in-context examples and input claim. In contrast to existing prompting methods, in \ours, LLM receives the task description via in-context examples. As a result, the performance of fact verification is highly sensitive to the selection of $\{ (x_i, y_i) \}_{i=1}^{k}$. To this end, we design a simple and efficient example retrieval pipeline, which is designed to choose semantically similar examples from the training set to maximize the relevance and informativeness of demonstrations $\{ (x_i, y_i) \}_{i=1}^{k}$ during in-context learning.

Specifically, we adopt a pretrained embedding model, denoted with $f_{\mathrm{embed}}$ (kept frozen in our \ours framework). The objective of our retrieval pipeline is to identify a set of $k$ examples $\{ (x_i, y_i) \}_{i=1}^{k}$ for each claim $x$, with:
\begin{equation}
    \begin{aligned}
    \{ (x_i, y_i) \}_{i=1}^{k} = \mathrm{topk}(\{ \mathrm{sim}(&f_{\mathrm{embed}}(x), \\ &f_{\mathrm{embed}}(x_i)) \}_{i=1}^{|\mathcal{X}^{\mathrm{train}}|}),
    \end{aligned}
    \label{eq:icl-retrieval}
\end{equation}
where topk returns $k$ largest elements from the given set (i.e., claims with highest similarity to $x$), while sim represents the cosine similarity function (i.e., $\mathrm{sim}(a, b) = \frac{a \cdot b}{\| a \| \| b \|}$). In essence, \Cref{eq:icl-retrieval} encodes the examples from the training set $\mathcal{X}^{\mathrm{train}}$ (only needs to be performed once), and then identifies the top-$k$ nearest elements by computing the highest cosine similarity scores. Overall, our in-context example retrieval pipeline performs similarity-based filtering to select semantically relevant examples, and thus optimizes the prior distribution for in-context learning. We additionally apply similarity thresholding by establishing a minimum cosine similarity of $0.5$, and set $k=10$ as the maximum number of demonstrations. In our implementation, SimCSE-RoBERTa is employed as the embedding function $f_{\mathrm{embed}}$ to encode input claims~\cite{liu2019roberta, gao-etal-2021-simcse}.

\begin{figure*}[t]
    \centering
    \includegraphics[trim=1.1cm 1.8cm 1.1cm 1.8cm, clip, width=1.0\linewidth]{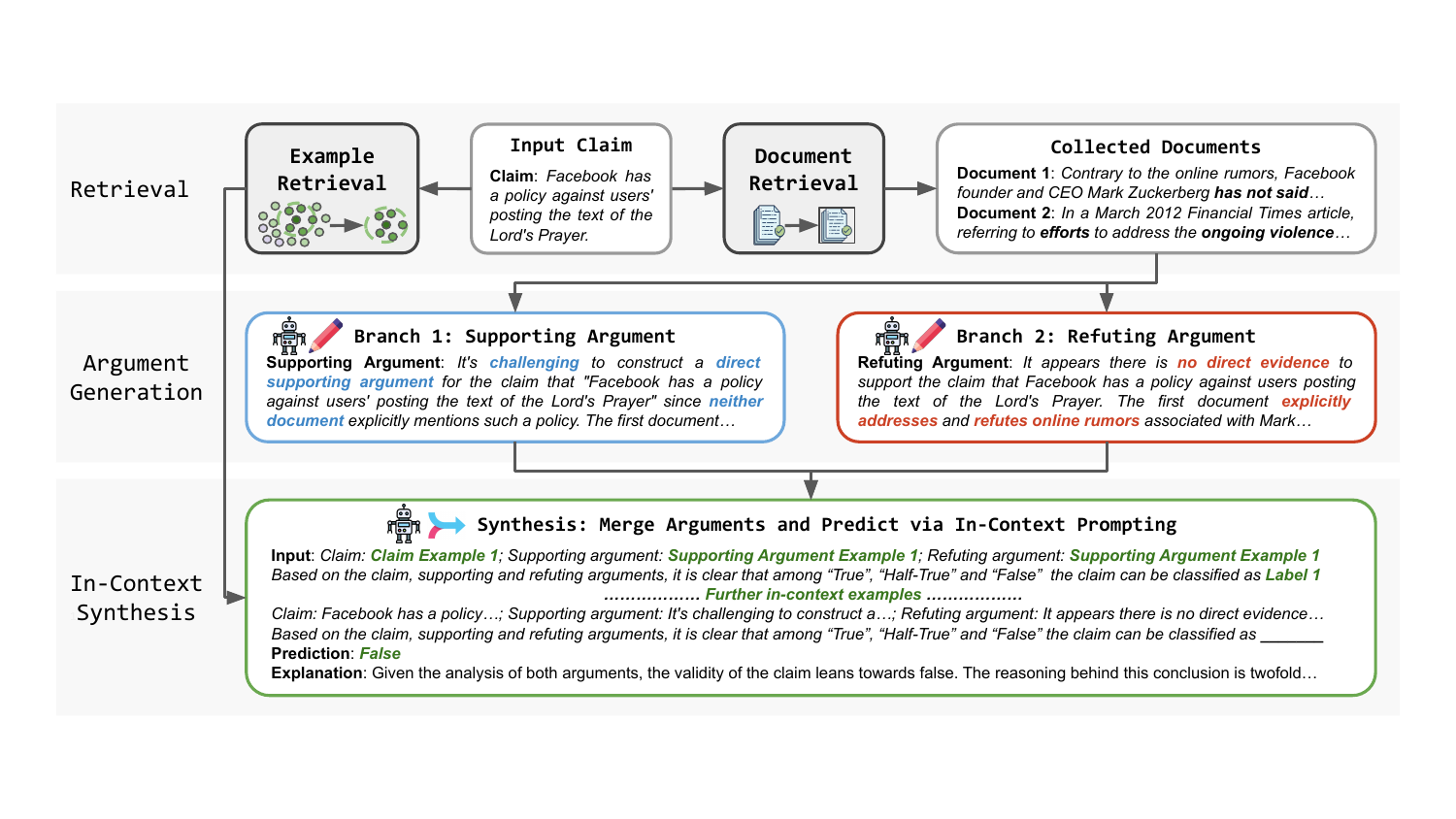}
    \caption{The proposed \ours, which performs few-shot fact verification by incorporating informative in-context demonstrations and contrastive arguments with nuanced information derived from the retrieved documents.}
    \label{fig:method}
\end{figure*}

\subsection{Document Retrieval}
\label{sec:retrieval}
The majority of RAG and fact-checking methods utilize sparse retrieval algorithms, dense retrieval models or third-party APIs to collect relevant documents~\cite{izacard-grave-2021-leveraging, izacard2022few, ram2023context, wang-shu-2023-explainable, zhang-gao-2023-towards}. While sparse retrieval methods are widely used, they often fall short in delivering optimal retrieval results for knowledge-intensive tasks like fact verification. On the other hand, dense retrieval methods suffer from efficiency issues in processing massive document collections and require extensive annotated data for optimal performance. These constraints render current retrieval approaches less effective for fact verification, where no~/~limited annotated claim-document pairs are available for training.

Therefore, we propose a two-stage pipeline $f_{\mathrm{retrieve}}$ in \ours that performs coarse-to-fine retrieval, which improves both computation efficiency and retrieval performance. Specifically, our pipeline includes: (1)~sparse retrieval via BM25, which collects a subset $\{ d_i \}_{i=1}^{\hat{m}}$ from a large collection of documents; and (2)~an dense retrieval model (denoted with $\theta$) that re-ranks and refines the selection of retrieved documents. Based on $x$, the first step narrows down to a subset from a much larger collection $\{ d_i \}_{i=1}^{n}$, while the learnable dense retriever further selects the $m$ most relevant documents $\{ d_i \}_{i=1}^{m}$ to verify input validity. Although BM25 may retrieve less relevant or even irrelevant elements, we note that with proper selection of $\hat{m}$, the desired documents tend to be found within the retrieved set for most cases. In our implementation, we use $\hat{m} = 20$ and $m = 5$ to balance the document retrieval performance and efficiency.

Following the sparse BM25 retrieval, we elaborate the learning of our dense retrieval model. To enhance re-ranking performance with limited annotated data, we exploit the BM25 scores as a coarse estimation of claim-document relevance. That is, we utilize the BM25 scores from the previous retrieval stage, combined with a limited collection of annotated examples, to train the dense retriever model. Specifically for claim-document pair $(x, d)$, we sample $l$ positive documents $\{ d_i^p \}_{i=1}^{l}$ and $l$ negative documents $\{ d_i^n \}_{i=1}^{l}$ based on the BM25 and inverse BM25 scores, which avoids introducing extensive noise in training. Using the sampled documents, we construct a ranking loss to expand the margin between document $d$ and the highest ranked document from $\{ d_i^p \}_{i=1}^{l}$ (i.e., {\small $f_{\mathrm{den}}(x, d) - \max(\{ f_{\mathrm{den}}(x, d_i^p) \}_{i=1}^{l})$}). In addition, we enhance the relevance between input-document pairs by imposing a penalty when the margin is below threshold $\tau$. Furthermore, our training objective incorporates a contrastive term derived from InfoNCE~\cite{chen2020simple, yue2024evidence}, which improves the relevance estimation between input-document pairs by `pushing away' negative documents. Overall, the optimization objective is:
\begin{equation}
    \begin{aligned}
        \mathbb{E}_{(x, d) \sim \mathcal{X}} [ & \max (0, \max (\{ f(x, d_i^p) \}_{i=1}^{l}) - f(x, d) + \tau) \\
        & - \lambda \frac{\exp(f(x, d))}{\exp(f(x, d)) + \sum_{2l} \exp(f(x, d_i))} ],
    \label{eq:retriever-loss}
    \end{aligned}
\end{equation}
where $\sum_{2l} \exp(f(x, d_i))$ represents the exponential sum of both positive examples $\{ d_i^p \}_{i=1}^{l}$ and negative examples $\{ d_i^n \}_{i=1}^{l}$. $\tau$ is the ranking margin threshold and $\lambda$ is a scaling factor. For each pair of $x$ and $d$, the first term in \Cref{eq:retriever-loss} becomes relevant when $f_{\mathrm{den}}(x, d)$ does not exceed the highest ranked document (i.e., also known as hard negative) score by $\tau$. Moreover, the contrastive term maximizes exponential score of the input-document pair in contrast to the sum of scores from the sampled documents. Hence, the dense retriever model learns to prioritize highly relevant documents while effectively filtering out those of less relevance to improve fact verification performance.

\begin{table*}[t]
\small
\centering
\begin{tabular}{@{}lcccccccccc@{}}
\toprule
\multirow{2}{*}{\textbf{Model}} & \multicolumn{5}{c}{\textbf{MS MARCO}}                         & \multicolumn{5}{c}{\textbf{Check-COVID}}                                      \\ \cmidrule(l){2-6} \cmidrule(l){7-11} 
      & N@1 $\uparrow$ & N@3 $\uparrow$ & R@3 $\uparrow$ & N@5 $\uparrow$ & R@5 $\uparrow$ & N@1 $\uparrow$ & N@3 $\uparrow$ & R@3 $\uparrow$ & N@5 $\uparrow$ & R@5 $\uparrow$ \\ \midrule
TFIDF & 0.419          & 0.531          & 0.613          & 0.562          & 0.687          & 0.266          & 0.363          & 0.427          & 0.385          & 0.480          \\
BM25  & 0.665          & 0.746          & 0.801          & 0.760          & 0.836          & 0.292          & 0.395          & 0.467          & 0.426          & 0.545          \\
DPR   & 0.738          & 0.793          & 0.850          & 0.797          & 0.903          & 0.324          & 0.411          & 0.477          & 0.457          & 0.588          \\
E5    & \ul{0.796}     & \ul{0.855}     & \ul{0.895}     & \ul{0.865}     & \ul{0.920}     & \ul{0.445}     & \ul{0.584}     & \ul{0.679}     & \ul{0.609}     & \ul{0.741}     \\ \midrule
\ours & \textbf{0.802} & \textbf{0.858} & \textbf{0.896} & \textbf{0.868} & \textbf{0.920} & \textbf{0.513} & \textbf{0.631} & \textbf{0.712} & \textbf{0.646} & \textbf{0.750} \\ \bottomrule
\end{tabular}
\caption{Evaluation results on document retrieval, with best results in bold and second best results underlined.}
\label{tab:retrieval}
\end{table*}

\subsection{Fact Verification by Synthesizing Contrastive Arguments}
To facilitate fact verification with LLMs, existing methods leverage intricate templates and techniques such as chain-of-thought (CoT), which decomposes input claims into sub-claims to verify~\cite{wei2022chain, wang-shu-2023-explainable}. Yet when assessing the (sub)-claims, current approaches prompt LLMs to perform binary classification (i.e., true or false), and thus often fail to incorporate nuanced information for fine-grained fact-checking~\cite{zhang-gao-2023-towards, pelrine-etal-2023-towards}. Moreover, the extended context created by retrieved demonstrations and documents can impair performance in LLMs with limited context windows or in smaller LLMs. Therefore, we propose a branching approach by generating and synthesizing contrastive arguments, in which we: (1)~decompose the fact-checking task into generating supporting and refuting arguments upon input claim and retrieved documents; and (2)~learn from informative in-context examples to synthesize the contrasting arguments, which incorporates adaptive prior knowledge and varying viewpoints. 

Provided with claim and retrieved documents, our first sub-task involves creating two branches in parallel that generate independent yet varying arguments from two opposing perspectives. In particular, we leverage the text comprehension and summarization capabilities of LLMs and perform instruction prompting to extract relevant facts and generate supporting~/~refuting arguments. We adopt a simple task description and optimize it to obtain concise, yet accurate arguments within a few sentences. For input $x$ and retrieved documents $\{ d_i \}_{i=1}^{m}$, the generated supporting~/~refuting arguments are denoted with $s$ and $r$. Therefore, for a specific example $(x, y) \sim \mathcal{X}$, we enrich the input to $(s, r, x, y)$ by integrating both supporting and refuting arguments. Notably, if no pertinent evidence is found to form the argument, LLMs are instructed to recognize the absence of evidence, as illustrated in branch 1 of \Cref{fig:method}. Consequently, this allows us to guide LLMs to take both arguments into consideration, facilitating a comprehensive analysis on the claim and its credibility.

Moving to the argument synthesis and inference phase (i.e., in-context synthesis) of our fact-checking framework, we aim to generate accurate prediction on the claim validity by leveraging in-context examples along with the contrasting arguments. Recall our in-context learning framework condition on the $k$-shot examples $\{ (x_i, y_i) \}_{i=1}^{k}$, we also incorporate the generated arguments and reformulate our inference with:
\begin{equation}
    \hat{y} = \arg\max_{y} f_{\mathrm{check}}(y | \{ s_i, r_i, x_i, y_i \}_{i=1}^{k}, s, r, x),
\end{equation}
where $s_i$, $r_i$ are the supporting and refuting arguments for the $i$-th demonstration. Note that the documents $\{ d_i \}_{i=1}^{m}$ are implicitly included in the arguments and thus no longer used in the prompt. At this point, we adopt the following template for each example in the final prompt:
\begin{displayquote}
Claim: \texttt{Claim}\\
Supporting argument: \texttt{Supporting Arg}\\
Refuting argument: \texttt{Refuting Arg}\\
Based on the claim, its supporting and refuting arguments, it is clear that among \texttt{Classes}, the claim should be classified as \texttt{Label}.
\end{displayquote}
Here, \texttt{Claim}, \texttt{Supporting Arg}, \texttt{Refuting Arg}, \texttt{Classes} are populated with the input claim, supporting and refuting arguments and the set of all classes. For in-context examples, \texttt{Label} is filled with the respective example's label, whereas for the target example, \texttt{Label} is left blank for prediction. Following the prediction, the explanation is generated in a similar fashion by integrating both arguments and prompting with instruction. 

\subsection{Summary of \ours}
Overall, the proposed \ours has three components: (1)~example retrieval; (2)~document retrieval; and (3)~in-context fact verification. The first two components are designed to collect relevant demonstrations and supporting documents that provide insightful context information. In the third component, we propose to generate contrasting arguments upon the retrieved documents, followed by incorporating these perspectives in inference to achieve fine-grained fact verification. With informative in-context examples featuring contrastive arguments, \ours can perform well regardless of the LLM size. To demonstrate the efficacy of \ours, we perform extensive experiments on multiple fact verification datasets, revealing that \ours can surpass state-of-the-art fact-checking methods even with a significantly smaller LLM.

\begin{table*}[t]
\small
\centering
\begin{tabular}{@{}llccccccccccc@{}}
\toprule
\multirow{2}{*}{\textbf{Model}} & \textbf{} & \multicolumn{3}{c}{LIAR}                         & \textbf{} & \multicolumn{3}{c}{RAWFC}                        &  & \multicolumn{3}{c}{ANTiVax}                      \\ \cmidrule{3-5} \cmidrule{7-9} \cmidrule{11-13} 
                                &           & P $\uparrow$   & R $\uparrow$   & F1 $\uparrow$  &           & P $\uparrow$   & R $\uparrow$   & F1 $\uparrow$  &  & P $\uparrow$   & R $\uparrow$   & F1 $\uparrow$  \\ \midrule
dEFEND                          &           & 0.230          & 0.185          & 0.205          &           & 0.449          & 0.432          & 0.440          &  & 0.729          & 0.839          & 0.781          \\
SentHAN                         &           & 0.226          & 0.200          & 0.212          &           & 0.457          & 0.455          & 0.456          &  & 0.691          & \textbf{0.984} & 0.812          \\
SBERT-FC                        &           & 0.241          & 0.221          & 0.231          &           & 0.511          & 0.460          & 0.484          &  & 0.736          & 0.951          & 0.830          \\
CofCED                          &           & 0.295          & 0.296          & 0.295          &           & 0.530          & 0.510          & 0.520          &  & 0.731          & \ul{0.956}    & 0.828          \\ \midrule
GPT-3.5                          &           & 0.291          & 0.251          & 0.270          &           & 0.485          & 0.485          & 0.485          &  & 0.771          & 0.850          & 0.808          \\
CoT                             &           & 0.226          & 0.242          & 0.237          &           & 0.424          & 0.466          & 0.444          &  & 0.816          & 0.877          & 0.845          \\
ReAct                           &           & 0.332          & 0.290          & 0.310          &           & 0.512          & 0.485          & 0.498          &  & 0.820          & 0.864          & 0.841          \\
HiSS                            &           & 0.468          & \ul{0.313}    & 0.375          &           & 0.534          & \textbf{0.544} & 0.539          &  & 0.823          & 0.887          & 0.853          \\ \midrule
\ours (w/ Mistral 7B)           &           & \textbf{0.616} & 0.305          & \ul{0.408}    &           & \ul{0.626}    & 0.516          & \ul{0.566}    &  & \ul{0.839}    & 0.873          & \ul{0.854}    \\
\ours (w/ GPT-3.5)               &           & \ul{0.471}    & \textbf{0.379} & \textbf{0.420} &           & \textbf{0.628} & \ul{0.526}    & \textbf{0.573} &  & \textbf{0.886} & 0.908          & \textbf{0.897} \\ \bottomrule
\end{tabular}
\caption{Evaluation results on fact verification, with best results in bold and second best results underlined.}
\label{tab:main-results}
\end{table*}
\section{Experiments}
\label{sec:exp}

\subsection{Experiment Design}
\textbf{Document Retrieval.} Our example retrieval model $f_{\mathrm{embed}}$ uses the pretrained SimCSE-RoBERTa~\cite{liu2019roberta, gao-etal-2021-simcse}. The document retrieval model $f_{\mathrm{retriever}}$ consists of BM25 and a dense retriever initialized with E5 (base)~\cite{wang2022text}. We adopt MS MARCO and Check-COVID dataset for document retrieval~\cite{nguyen2016ms, wang-etal-2023-check-covid}. The adopted metrics are NDCG and Recall (i.e., N@$k$ and R@$k$) with $k \in [1, 3, 5]$. For baselines, we adopt the sparse TFIDF and BM25 and dense models DPR and E5~\cite{karpukhin-etal-2020-dense, wang2022text}.

\noindent
\textbf{Fact Verification.} We adopt Mistral 7B and GPT-3.5 as our base LLM~\cite{jiang2023mistral, ouyang2022training}. We adopt three datasets with varying granularity: LIAR (True~/~Mostly-true~/~Half-true~/~Barely-true~/~False~/~Pants-fire)), RAWFC (True~/~Half-true~/~False), and ANTiVax (True~/~False)~\cite{wang-2017-liar, yang-etal-2022-coarse, hayawi2022anti}. For LIAR and RAWFC, we adopt Wikipedia as document sources and use the MS MARCO trained retriever. The document collection for ANTiVax is collected from CORD and LitCOVID, thus we use the Check-COVID trained retriever~\cite{karpukhin-etal-2020-dense, wang2020cord, chen2021litcovid}. Our supervised baselines are dEFEND, SentHAN, SBERT-FC and CofCED~\cite{shu2019defend, ma-etal-2019-sentence, kotonya-toni-2020-explainable-automated, yang-etal-2022-coarse}. GPT-3.5-based methods include GPT-3.5, CoT, ReAct and HiSS~\cite{brown2020language, wei2022chain, yao2022react, zhang-gao-2023-towards}. We adopt macro recall, precision and F1 scores to evaluate fact-checking performance. Automated evaluation is used for explanation quality, including politeness, factuality and claim-relevance following~\cite{he2023reinforcement}.

\subsection{Document Retrieval}
Our document retrieval results are reported in \Cref{tab:retrieval}. In this table, rows represent retrieval methods and the columns represent different datasets~/~metrics. For top-1 scores, we use N@1 since top-1 NDCG and Recall scores are equivalent in this case. From the results we observe: (1)~\ours retriever consistently outperforms baseline methods across all metrics, with an average performance improvement of $3.56\%$ across metrics and datasets. (2)~In contrast to sparse retrieval along, the additional dense retriever significantly improves the ranking performance. For example, \ours achieves $37.61\%$ performance improvement in Recall@5 compared to BM25 on Check-COVID. (3)~The performance gains through our retrieval pipeline are more significant on the Check-COVID dataset. For instance, the relative improvement of NDCG@5 shifts from $0.35\%$ to $8.05\%$ when moving from MS MARCO to Check-COVID. Overall, we find the proposed retrieval pipeline in \ours performs well in collecting relevant documents. In addition, the retrieval pipeline proves to be essential for specialized domains like healthcare (e.g., COVID), leading to notable performance improvements.

\subsection{Fact Verification}
We proceed to discuss our fact verification performance of \ours, with the results reported in \Cref{tab:main-results}. Similarly, methods are depicted in rows and datasets~/~metrics are represented in columns. The first group of baseline methods comprise supervised approaches (i.e. from dEFEND to CofCED), followed by methods built upon GPT-3.5 (i.e. from GPT-3.5 to HiSS), and the bottom row incorporate \ours with Mistral 7B and GPT-3.5. We use P, R and F1 to abbreviate precision, recall and F1 scores\footnote{In our experiments, F1 score is favored as it balances the trade-off between precision and recall, thereby offering a more comprehensive performance measure for fact verification.}, and we observe: (1)~Both \ours variants demonstrates superior fact-checking performance across all datasets. For example, \ours with Mistral 7B outperforms the best baseline method in F1 by $8.8\%$, while \ours with GPT-3.5 achieves a significant $12.0\%$ performance gain on F1. (2)~\ours with GPT-3.5 delivers the best classification results overall. In particular, it leads in precision~/~recall on two of the three datasets and achieves the highest F1 for all datasets, averaging a $7.8\%$ increase in F1 performance. (3)~Notably, \ours w/ Mistral 7B backbone is superior than all baseline methods on F1 scores despite its significantly smaller size (7B) than GPT-3.5. This suggests that the proposed in-context synthesis can extract concise yet informative arguments and help LLMs generate accurate predictions on claim credibility. In summary, the \ours can outperform state-of-the-art fact verification methods by a substantial margin. Even when utilizing a notably smaller model (Mistral 7B), \ours consistently exhibits superior performance, highlighting its efficacy in fact verification.

\subsection{Explanation Generation}
Based on the fact verification results, the explanations for the prediction can be generated in a similar fashion. To evaluate explanation quality, we benchmark against GPT-3.5 and HiSS, as supervised and the rest LLM methods are not designed to generate fact-checking explanations. We report the explanation quality results in \Cref{tab:explanation-results}, with Po., Fa. and Rel. representing politeness, factuality and claim relevance. For \ours, we use (M) and (G) to denote the Mistral 7B and GPT-3.5 backbones. Our findings are: (1)~both baselines and \ours perform well in generating explanations based on the fact-checking predictions, achieving average scores above 0.9 for both politeness and factuality. (2)~GPT-3.5-based methods show similar performance regardless of prompting strategies. For instance, the average scores on ANTiVax across metrics are 0.898, 0.909 and 0.915 for GPT-3.5, HiSS and \ours (G). (3)~Surprisingly, \ours with Mistral excels in explanation generation, achieving the highest politeness and factuality scores on all datasets, which may be attributed to the instruction-following capabilities of Mistral 7B. In sum, the explanation evaluation shows that \ours can consistently generate high-quality explanations regardless of the choice of the LLM.

\begin{table}[t]
\small
\centering
\begin{tabular}{@{}llccc@{}}
\toprule
\textbf{Dataset}                  & \textbf{Method} & \textbf{Po.} $\uparrow$ & \textbf{Fa.} $\uparrow$ & \textbf{Rel.} $\uparrow$ \\ \midrule 
\multirow{4}{*}{\textbf{LIAR}}    & GPT-3.5          & 0.947                 & 0.943                 & 0.846                    \\
                                  & HiSS            & 0.967                 & 0.964                 & 0.848                    \\
                                  & \ours (M)       & \textbf{0.973}        & \textbf{0.969}        & \textbf{0.883}           \\
                                  & \ours (G)       & \ul{0.969}            & \ul{0.969}            & \ul{0.852}               \\ \midrule
\multirow{4}{*}{\textbf{RAWFC}}   & GPT-3.5          & 0.965                 & 0.949                 & 0.856                    \\
                                  & HiSS            & \ul{0.971}            & 0.955                 & \ul{0.861}               \\
                                  & \ours (M)       & \textbf{0.974}        & \textbf{0.971}        & 0.757                    \\ 
                                  & \ours (G)       & 0.970                 & \ul{0.960}            & \textbf{0.862}           \\ \midrule
\multirow{4}{*}{\textbf{ANTiVax}} & GPT-3.5          & 0.958                 & 0.963                 & 0.774                   \\
                                  & HiSS            & 0.986                 & \ul{0.974}            & 0.768                   \\
                                  & \ours (M)       & \textbf{0.987}        & \textbf{0.976}        & \textbf{0.800}          \\ 
                                  & \ours (G)       & \ul{0.986}            & 0.973                 & \ul{0.785}              \\ \bottomrule
\end{tabular}
\caption{Evaluation results on explanation quality, with best results in bold and second best results underlined.}
\label{tab:explanation-results}
\end{table}

\section{Conclusion}
In this paper, we propose \ours, a novel retrieval augmented fact verification framework. \ours consists of three key components: (1)~example retrieval, which provides informative in-context demonstrations; (2)~document retrieval that collects relevant documents from verifiable sources; and (3)~in-context prompting, where few-shot fact-checking is performed by considering both informative examples and nuanced information from contrastive arguments. As a result, \ours achieves fine-grained fact verification without the need for complex prompting techniques and large-size LLMs. Our experiment results on benchmark datasets highlight the superiority of \ours, which consistently outperforms state-of-the-art methods methods in both fact-checking performance and the quality of generated explanations.

\section{Limitations}
Despite introducing \ours for retrieval augmented fact verification, we have not discussed the setting in which the document retrieval domain significantly differs from the fact-checking domain (e.g., using Wikipedia documents to fact-check COVID misinformation), which can cause performance deterioration for domain-generalized applications. Furthermore, we have not examined the robustness and reliability of our example retrieval and document retrieval, which could unlock additional improvements for fact verification. Consequently, we plan to explore a more generalized and domain-adaptive solution for retrieval augmented fact verification as future work.


\bibliography{anthology,custom}
\bibliographystyle{acl_natbib}

\end{document}